%% file: acl_latex.tex
\pdfoutput=1

\documentclass[11pt]{article}

\usepackage[final]{acl}

\usepackage{times}
\usepackage{latexsym}

\usepackage[T1]{fontenc}

\usepackage[utf8]{inputenc}

\usepackage{microtype}

\usepackage{inconsolata}

\usepackage{amsmath}
\usepackage{amsfonts}
\usepackage{amssymb}
\usepackage{subfigure}
\usepackage{cases}
\usepackage{dsfont}
\usepackage{bm}
\usepackage{bbm}
\usepackage{graphicx}
\usepackage{color}
\usepackage{multicol}
\usepackage{multirow}
\usepackage{wrapfig,lipsum,booktabs}
\usepackage[ruled,vlined,linesnumbered]{algorithm2e}
\usepackage{pgfplots}
\pgfplotsset{compat=1.12} 
\usepackage{filecontents}
\usepackage{tikz}
\usepackage{xcolor}
\usepackage{colortbl}
\usepackage{xspace}
\usepackage{makecell}
\usepackage{soul}
\usepackage{subcaption}
\usetikzlibrary{calc}
\usepgfplotslibrary{groupplots}
\usetikzlibrary{angles,quotes} 
\usetikzlibrary{shapes,arrows}
\usetikzlibrary{backgrounds}
\usetikzlibrary{matrix}
\usetikzlibrary{patterns}
\usepackage{tikz-3dplot}
\usepackage{hyperref}
\usepackage{cleveref}
\usepackage{paralist}
\usepackage{cancel}
\usepackage{todonotes}
\usepackage{tabu}
\usepackage{rotating}
\usepackage{etoolbox}
\usepackage{adjustbox}
\usepackage{enumerate}
\usepackage{enumitem}
\setitemize{itemsep=0pt,topsep=0pt,parsep=0pt,partopsep=0pt}
\setenumerate{itemsep=0pt,topsep=0pt,parsep=0pt,partopsep=0pt}
\usepackage{pifont}
\usepackage{cancel}
\usepackage{lipsum}
\usepackage{listings,lstautogobble}
\usepackage{fancyvrb}
\usepackage{fvextra}
\usepackage{caption}
\usepackage{arydshln}
\usepackage{float}
\usepackage{lipsum} 
\usepackage[dvipsnames]{xcolor}
\newcommand{\model}[1]{\textsc{#1}\xspace}

\newcommand{\baseline}{\model{Baseline}}
\newcommand{\qwen}{\model{Qwen2.5-Math-7B-Instruct}}
\newcommand{\llama}{\model{Llama-3.1-8B-math}}
\newcommand{\mistral}{\model{MetaMath-Mistral-7B}}
\newcommand{\deepseek}{\model{DeepSeekMath-7B-Instruct}}
\newcommand{\en}{\model{PRM-cross}}
\newcommand{\mono}{\model{PRM-mono}}
\newcommand{\mix}{\model{PRM-multi}}
\newcommand{\scmethod}{\model{SC}}
\newcommand{\orm}{\model{ORM}}
\newcommand{\prm}{\model{PRM}}

\newcommand{\dataset}[1]{\texttt{#1}\xspace}
\newcommand{\mathset}{\dataset{MATH500}}
\newcommand{\mgsmset}{\dataset{MGSM}}

\newcommand{\avgall}{\mu_{\textsc{all}}}
\newcommand{\avgseen}{\mu_{\textsc{seen}}}
\newcommand{\avgunseen}{\mu_{\textsc{unseen}}}

\newcommand{\newrevision}{\textcolor{black}}

\title{Demystifying Multilingual Chain-of-Thought in Process Reward Modeling}

\author{%
Weixuan Wang\textsuperscript{1} \quad Minghao Wu\textsuperscript{2} \quad Barry Haddow\textsuperscript{1} \quad Alexandra Birch\textsuperscript{1} \\[1ex]
\textsuperscript{1}School of Informatics, University of Edinburgh \\
\textsuperscript{2}Monash University \\
\texttt{\{weixuan.wang, bhaddow, a.birch\}@ed.ac.uk} \\
\texttt{minghao.wu@monash.edu} 
}

\begin{document}
\maketitle

\begin{abstract}

Large language models (LLMs) are designed to perform a wide range of tasks. To improve their ability to solve complex problems requiring multi-step reasoning, recent research leverages process reward modeling to provide fine-grained feedback at each step of the reasoning process for reinforcement learning (RL), but it predominantly focuses on English. In this paper, we tackle the critical challenge of extending process reward models (PRMs) to multilingual settings. To achieve this, we train multilingual PRMs on a dataset spanning seven languages, which is translated from English. Through comprehensive evaluations on two widely used reasoning benchmarks across 11 languages, we demonstrate that multilingual PRMs not only improve average accuracy but also reduce early-stage reasoning errors. Furthermore, our results highlight the sensitivity of multilingual PRMs to both the number of training languages and the volume of English data, while also uncovering the benefits arising from more candidate responses and trainable parameters. This work opens promising avenues for robust multilingual applications in complex, multi-step reasoning tasks. In addition, we release the code to foster research along this line.\footnote{\url{https://github.com/weixuan-wang123/Multilingual-PRM}}

\end{abstract}

\input{1_introduction}

\input{2_related_work}

\input{3_method}
\input{4_setting}

\input{5_results}

\input{6_analysis}
\input{7_conclusion}
\input{8_limitations}
\input{10_ackowledgement}

\bibliography{custom}

\clearpage
\appendix

\input{9_appendix}

\end{document}

%% file: 1_introduction.tex
\section{Introduction}
\label{sec:intro}

Aligning large language models (LLMs) with human preferences can significantly improve the model performance across various downstream tasks \citep{christiano2017deep,ziegler2019fine}. This requires a reward model that is trained on human preference data \citep{ziegler2019fine,stiennon2020learning,shen2021generate,ouyang2022training}. Typically, reward models are trained based on the final outcome of the LLMs' response, and we refer to these as outcome reward models (ORMs) \citep{orm1,solving,orm2}. However, most of recent work demonstrates that ORMs fall short on complex multi-step reasoning tasks \citep{solving,deepseek}. To overcome this limitation, process reward models (PRMs) are introduced, providing fine-grained rewards at each step of the LLMs' reasoning \citep{prm800k,making,shepherd,let}. Previous research has shown that LLMs supervised by PRMs can effectively produce better responses \citep{shepherd,deepseek}.

Despite these significant advances, recent research on ORMs and PRMs has predominantly focused on monolingual settings, particularly English \citep{prm800k,openr,shepherd}. However, the exploration of multilingual PRMs remains relatively limited. Therefore, with the advent of multilingual LLMs, a natural research question arises: \textit{How can we effectively train multilingual PRMs for complex, multi-step reasoning tasks?}

To address this research question, we translate the existing PRM datasets, PRM800K \citep{prm800k} and Math-Shepherd \citep{shepherd}, from English into six additional languages, resulting in a total of seven seen languages for training. We then train multilingual PRMs using the collection of these translated datasets. We define three PRM setups: \mono, \en, and \mix. The \mono setup is trained and evaluated solely on a single language, the \en setup is trained on one language but evaluated on all test languages, and the \mix setup is trained on seven seen languages and evaluated on all test languages. Finally, we conduct a comprehensive evaluation on two popular reasoning tasks (\mathset and \mgsmset) across 11 languages (seven seen languages and four unseen languages) using three LLMs (\mistral, \llama, and \deepseek).

Our main takeaways are summarized as follows:
\begin{itemize}

    \item \textbf{Multilingual PRM consistently outperforms monolingual and cross-lingual PRMs across all three LLMs.} Our results demonstrate that \mix significantly improves model performance, boosting average accuracy by up to +1.2 and +1.5 points compared to \en and \mono, respectively (see \autoref{sec:results_multi_mono}).

    \item \textbf{Multilingual PRM is sensitive to both the number of languages and the amount of English training data.} Our experiment shows that training an optimal multilingual PRM requires careful consideration of how many languages to include (see \autoref{sec:results_languages_number}) and how much English data to use (see \autoref{sec:results_english_percentage}).

    \item \textbf{Multilingual PRM produces fewer errors in the early steps.} We identify the first occurrences of wrong predictions made by PRMs and observe that \mix produces fewer errors in the early steps compared to \mono and \en (see \autoref{sec:analysis_error}).

    \item \textbf{Multilingual PRM can benefit more from more candidate responses and trainable parameters.} Our analysis demonstrates that \mix becomes more advantageous with a larger number of candidate responses (see \autoref{sec:analysis_samplecounts}) and when more trainable parameters are introduced (see \autoref{sec:analysis_lora}).

\end{itemize}

%% file: 2_related_work.tex
\section{Related Work}
\label{sec:related_work}

\paragraph{Reward Model in Mathematical Reasoning}

To advance the accuracy of mathematical reasoning, reward models (RMs) have emerged as powerful tools for evaluating and guiding solution generation. In particular, two principal RM paradigms have garnered significant attention: the Outcome Reward Models (ORMs) \citep{orm1,orm2} and the Process Reward Models (PRMs) \citep{solving,prm800k,making,let,shepherd,improve,bugs,openr}. ORMs assign a single score to an entire solution and thereby focuses on final correctness, whereas PRMs score each individual step of the reasoning process, offering more finer-grained evaluations. As a result, PRMs provide more detailed guidance and have demonstrated greater potential in enhancing reasoning capabilities compared to ORMs \citep{prm800k,fine}.

\paragraph{Multilingual Reward Model}

Beyond English-language tasks, the integration of RMs into multilingual scenarios is still under-explored. 
Reinforcement learning approaches often rely on RMs predominantly trained on English data \citep{deepseek,qwen}. This over-representation introduces biases, as these RMs may overfit to English-specific syntactic and semantic patterns, limiting their effectiveness in cross-lingual tasks and motivating the development of multilingual RMs \citep{crossrm}. While there is growing evidence that cross-lingual transfer is feasible \citep{reuse,crossrm}, existing research often overlooks the unique challenges of multilingual reasoning.
After the release of the OpenAI-o1 model \citep{o1}, PRMs, with their capability for fine-grained feedback, have attracted even greater interest. Yet, the performance of multilingual PRMs in diverse linguistic contexts remains insufficiently investigated \citep{imbalance}. To bridge this gap, we investigate how multilingual PRMs contribute to solving mathematical tasks across different languages, aiming to provide insights into how fine-grained process supervision can enhance reasoning capabilities beyond English, thereby contributing to the development of more universally applicable reasoning models.

%% file: 3_method.tex
\begin{figure}[t]
    \centering
    \includegraphics[scale=0.3]{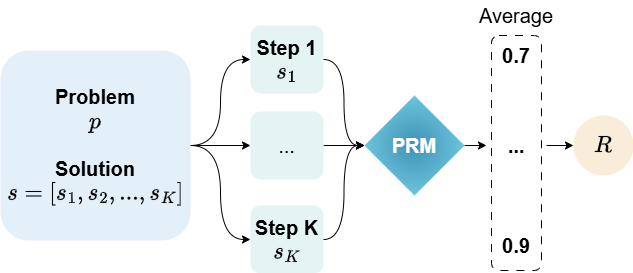}
    \caption{Framework of PRM.}
    \label{fig:PRM}
\end{figure}

\section{Process Reward Modeling}
\label{sec:prm}

\subsection{PRM Training}

Given a question $p$ and its solution $s$, the ORM assigns a single value to $s$ to indicate if $s$ is correct. 
We stack a binary classifier on top of the LLM and train the ORM with the binary cross-entropy loss:
\begin{equation}
\begin{aligned}
    \mathcal{L}_{\text{ORM}}& = \\
    & -{(y_s \log(r_s) + (1 - y_s)\log(1 - r_s))}
\end{aligned}
\end{equation}
where $y_s$ is the ground truth label for the solution $s$ ($y_s = 1$ if $s$ is correct, otherwise $y_s = 0$), and $r_s$ is the probability score that $s$ is correct.

In contrast, the PRM evaluates each reasoning step of the solution $s$. The PRM is trained using the following loss function:
\begin{equation}
\begin{aligned}
    \mathcal{L}_{\text{PRM}} & = \\
    -\sum_{i=1}^{K}& y_{s_i}\log(r_{s_i}) + (1 - y_{s_i})\log(1 - r_{s_i})
\end{aligned}
\end{equation}
where $s_i$ is the $i$-th step of the solution $s$, $y_{s_i}$ is the ground truth label for step $s_i$, $r_{s_i}$ is the score assigned to $s_i$ by the PRM, and $K$ is the total number of reasoning steps in the solution $s$. Compared to ORM, PRM provides more detailed and reliable feedback by evaluating individual steps.

\subsection{Ranking for Verification}
\label{sec:prm_verify}

Following \citet{multiorm,prm800k,shepherd}, we evaluate the performance of PRM using the \textit{best-of-N} selection evaluation paradigm \citep{bestofn1,bestofn2}. Specifically, given a question, multiple solutions are sampled from an LLM (referred to as the \textit{generator}) and re-ranked using a reward model (referred to as the \textit{verifier}). For each solution, as shown in \autoref{fig:PRM}, PRM assesses the correctness of each reasoning step. The scores for all steps are averaged to compute an overall score for the solution. The highest-scoring solution is then selected as the final output. This approach enhances the likelihood of selecting solutions containing correct answers, thereby improving the success rate of solving mathematical problems with LLMs.

\subsection{Reinforcement Learning with Process Supervision}

Using the trained PRM, we fine-tune LLMs with Policy Optimization (PPO) \citep{ppo} in a step-by-step manner. This method differs from the conventional strategy that uses PPO with an ORM, that only gives a reward at the end of the response. Conversely, our step-by-step PPO offers rewards at the end of each reasoning step. 

While we analyse PRM both intrinsically (using best-of-N), and extrinsically (using PPO), we focus on best-of-N for a clean testbed without confounders from reinforcement learning.

%% file: 4_setting.tex
\section{Experimental Setups}
\label{sec:setting}

\paragraph{Training Datasets}
We combine the PRM800K \citep{prm800k} and Math-Shepherd \citep{shepherd} as training data to finetune PRMs, and translate the combined dataset from English (en) to six languages: German (de), Spanish (es), French (fr), Russian (ru), Swahili (sw), and Chinese (zh) with using NLLB 3.3B \citep{nllb}. The reasoning step statistics are presented in \autoref{tab:data_stat} (\autoref{sec:appendix_statistics}), and 
the parallel examples across seven languages have the same number of reasoning steps. \newrevision{To ensure high translation quality, we use regular expressions to filter out translated training instances that contain discrepancies in numbers or equations compared to the original English dataset.}

\paragraph{Test Dataset}
We evaluate the performance of LLMs using two widely used math reasoning datasets, \mgsmset \citep{mgsm} and \mathset \citep{shepherd}. For the \mathset datset, we translate it from English to ten languages: Bengali (bn), German (de), Spanish (es), French (fr), Japanese (ja), Russian (ru), Swahili (sw), Telugu (te), Thai (th), and Chinese (zh) with Google Translate, which is consistent with the languages included in the \mgsmset dataset. 
Furthermore, we also categorize the languages involved in the downstream tasks into two groups based on the training data of \prm: \textit{seen languages} (en, de, es, fr, ru, sw, and zh) and \textit{unseen languages} (bn, ja, te, and th).
\newrevision{To ensure the quality of our testset, we employ two human translators to post-edit the translated examples for each high-resource language (de, es, fr, ru, zh, and ja) and leverage \model{gpt-4o} to revise the translations in low-resource languages (bn, sw, te, and th). More details are shown in \autoref{sec:appendix_translation}.}

\input{tables/mix-mono-en-math}

\paragraph{Multilingual PRM Setups}
To better understand PRMs in the context of multilingual research, we define three setups: \mono, \en, and \mix. The \mono setup is trained and evaluated on the same single language, serving as the baseline for monolingual PRMs. The \en setup is trained on one language but evaluated on all 11 test languages. Specifically, in this work, we train \en on the English PRM dataset unless otherwise specified. Finally, the \mix setup represents the multilingual PRM, which is both trained on all the seen languages and evaluated on all 11 test languages. To enhance the reliability and generalizability of our study, we train our multilingual PRM (\textbf{\textit{verifier}}) based on the \qwen \citep{qwen}, and leverage three diverse LLMs as the \textbf{\textit{generator}}: \mistral \citep{metamath}, \llama (fine-tuned with the MetaMath dataset \citep{llama}),\footnote{\url{https://huggingface.co/gohsyi/Meta-Llama-3.1-8B-sft-metamath}} and \deepseek \citep{deepseek}. 
The details of training these PRMs are presented in \autoref{sec:appendix_training}.

%% file: tables/mix-mono-en-math.tex
\begin{table*}[t] \small
\centering
\setlength{\tabcolsep}{5pt}
\begin{tabular}{lcccccccccccccc}
\toprule
\mathset & $\avgall$      & $\avgseen$     & $\avgunseen$   & en             & de             & es             & fr             & ru             & sw             & zh             & ja             & bn             & te             & th             \\ \midrule
\multicolumn{15}{c}{\mistral} \\ \midrule
\baseline	 & 22.1	 & 24.3	 & 18.2	 & 26.8	 & 26.2	 & 28.2	 & 25.4	 & 27.4	 & 13.4	 & 23.0	 & 25.0	 & 18.0	 & 10.6	 & 19.2 \\
\mono             & \multicolumn{1}{c}{-}         & 42.5      & \multicolumn{1}{c}{-}            & 49.0                  & 44.4                  & 45.8                  & 45.6                  & 46.0                  & 25.0                  & 41.8                  & \multicolumn{1}{c}{-}  & \multicolumn{1}{c}{-}  & \multicolumn{1}{c}{-}  & \multicolumn{1}{c}{-}          \\
\en               & 39.4     & 43.1      & 39.1        & 49.0                  & 45.4                  & 45.0                  & \textbf{46.8}         & \textbf{46.4}         & 25.2                  & \textbf{43.8}         & 43.6         & 31.4                  & \textbf{22.0}         & 34.6      \\
\mix              & \textbf{39.6}                & \textbf{43.1}                 & \textbf{39.4}                   & \textbf{50.2}         & \textbf{45.6}         & \textbf{47.4}         & 45.4                  & 45.2                  & \textbf{25.2}         & 42.8                  & \textbf{43.6}         & \textbf{32.6}         & 21.8                  & \textbf{35.2}                 \\ \midrule
\multicolumn{15}{c}{\llama}  \\ \midrule
\baseline	 & 22.1	 & 24.3	 & 18.1	 & 30.4	 & 22.4	 & 27.4	 & 25.4	 & 22.0	 & 15.4	 & 27.4	 & 20.0	 & 16.6	 & 16.0	 & 19.8 \\
\mono             & -         & 43.3      & -            & 49.0                  & 46.2                  & 45.8                  & 44.2                  & 45.8                  & 26.2         & 46.2         & \multicolumn{1}{c}{-}  & \multicolumn{1}{c}{-}  & \multicolumn{1}{c}{-}  & \multicolumn{1}{c}{-}          \\
\en               & 40.9     & 43.6      & 36.3        & 49.0         & 48.8                  & \textbf{46.6}                  & 44.8                  & 44.8                  & 26.0                  & 45.2                  & \textbf{43.0}                  & \textbf{36.0}                  & 28.2                  & 37.8      \\
\mix              & \textbf{41.7}                & \textbf{44.8}                 & \textbf{36.4}                   & \textbf{51.0}                  & \textbf{48.8}         & 45.8         & \textbf{46.0}         & \textbf{46.2}         & \textbf{28.4}                  & \textbf{47.2}                  & 42.0         & 34.6         & \textbf{30.2}         & \textbf{38.6}                 \\ \midrule
\multicolumn{15}{c}{\deepseek} \\ \midrule
\baseline	 & 26.4	 & 32.5	 & 15.7	 & 42.0	 & 35.6	 & 36.4	 & 35.0	 & 36.4	 & 9.6	 & 32.4	 & 33.2	 & 9.8	 & 4.6	 & 15.2 \\
\mono             & \multicolumn{1}{c}{-}          & 55.1      & \multicolumn{1}{c}{-}            & 63.0                  & 59.0                  & \textbf{60.4}         & 59.0                  & 60.2                  & 29.2                  & \textbf{55.0}         & \multicolumn{1}{c}{-}  & \multicolumn{1}{c}{-}  & \multicolumn{1}{c}{-}  & \multicolumn{1}{c}{\textbf{-}} \\
\en               & 50.2     & 54.9      & 41.9        & 62.4                  & \textbf{60.0}         & 59.8                  & \textbf{61.4}         & 57.4                  & 29.4                  & 54.0                  & 54.4                  & \textbf{38.2}         & 32.4                  & 42.6      \\
\mix              & \textbf{51.3}                & \textbf{55.6}                 & \textbf{43.7}                   & \textbf{63.8}         & 58.6                  & 60.2                  & 60.2                  & \textbf{61.4}         & \textbf{30.6}         & 54.2                  & \textbf{55.8}         & 38.0                  & \textbf{35.6}         & \textbf{45.4}        \\
\bottomrule
\end{tabular}
\caption{\label{tab:mix-mono-en_math} Different PRMs' best-of-N sampling (N = 64) performance on \mathset with the generator of \mistral, \llama, and \deepseek. $\avgall$, $\avgseen$, and $\avgunseen$ indicate the macro-average of results across all the languages, the seen languages, and the unseen languages, respectively.}
\end{table*}

%% file: 5_results.tex
\section{Recipes for Multilingual PRM Training}
\label{sec:results}

In this section, we conduct a series of experiments to investigate the performance of multilingual PRM. We examine how \mix compares to \mono and \en (\autoref{sec:results_multi_mono}), the impact of the number of training languages (\autoref{sec:results_languages_number}), and the effect of varying the proportion of English in the training data (\autoref{sec:results_english_percentage}).

\subsection{Monolingual, Cross-lingual, or Multilingual PRMs?}
\label{sec:results_multi_mono}

Building on \citet{multiorm}'s findings that cross-lingual ORMs outperform monolingual ones, we investigate the impact of multilingualism on PRMs. Specifically, we compare \mono, \en, and \mix to determine which setup offers best performance across languages.

\paragraph{Setup}

We include three setups in this work. The \mono is trained and evaluated on each individual language from the set of seen languages. The \en is trained exclusively on an English dataset and evaluated on all 11 test languages. Finally, the \mix is trained on all seen languages and tested on all 11 test languages.

\paragraph{Multilingual PRMs perform best, followed by cross-lingual PRMs, while monolingual PRMs achieve the worst performance, on the seen languages.}

As shown in \autoref{tab:mix-mono-en_math}, \mix consistently achieves the highest performance across multiple language generators on the seen languages, surpassing \mono and \en by +1.5 and +1.2 with \llama generator, respectively. This indicates that incorporating data from multiple languages for PRM training significantly enhances the model's ability  across different languages. When comparing \mono and \en, we observe that \en outperforms the \mono for the English-centric \mistral and \llama generators. 
We hypothesize that this advantage stems from the pre-training phase: these generators are predominantly trained on English data but have limited exposure to multilingual corpora. As a result, fine-tuning on English PRM data enhances the reasoning capabilities of PRMs, facilitating greater cross-lingual transfer.
More monolingual results are in \autoref{sec:appendix_crosslingualprm}.

\paragraph{Multilingual PRMs generalize better on the unseen languages.}
Both \en and \mix are evaluated on four additional unseen languages. As shown in \autoref{tab:mix-mono-en_math}, \mix demonstrates superior overall performance on the unseen languages in terms of $\avgunseen$. These results suggest that training PRMs on multilingual datasets can effectively enhance model generalization to the unseen languages. More results on general-purpose LLM are provided in \autoref{sec:appendix_general}.

In conclusion, these findings demonstrate that training a single multilingual PRM is an effective strategy for broad cross-lingual coverage, outperforming models trained either on a target language or on English alone. This outcome supports that \mix is particularly advantageous for expanding the capabilities of PRMs in multilingual settings. More results on \mgsmset are in \autoref{sec:appendix_mix_mono_en_mgsm}.

\subsection{Does More Languages Lead to Better Multilingual PRMs?}
\label{sec:results_languages_number}

\begin{figure}[t]
    \centering
    \includegraphics[scale=0.5]{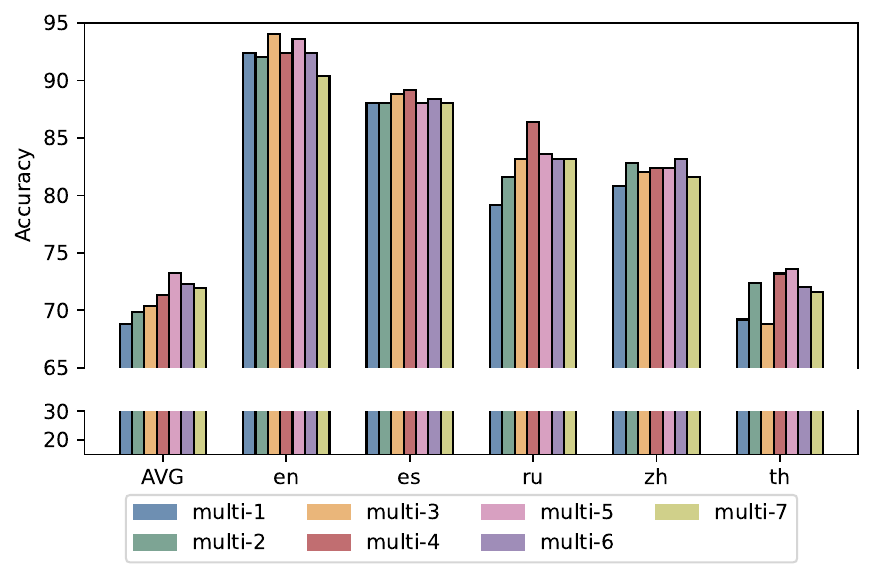}
    \caption{Best-of-N Performance on \mgsmset of PRMs trained using various subsets of English, German, Spanish, French, Russian, Swahili, and Chinese, with the generator of \llama. The averages scores across all 11 languages.}
    \label{fig:train-counts}
\end{figure}

While multilingual PRMs have demonstrated significant improvements, the question of how many languages are needed to achieve the best performance remains an open research problem. In this section, we address this research question by exploring the relationship between the number of training languages and the resulting performance.

\paragraph{Setup}
We conduct experiments by training PRMs on datasets ranging from a single language up to all seven languages. 

In this section, the number of total training examples of all PRMs are fixed.
When the number of languages exceeds one, the total training examples are evenly distributed across all the selected languages. For evaluation, we test all PRMs on 11 different languages. The evaluation scores are averaged for each test language across all PRMs trained with the same number of languages.

\paragraph{More languages do not result in better multilingual PRMs.}
As shown in \autoref{fig:train-counts}, the overall performance (\texttt{AVG}) improves as the number of training languages increases up to five languages. Beyond this point, adding more languages does not lead to further gains. Additionally, results from five individual languages (four seen languages and one unseen language) demonstrate that, although the optimal number of training languages varies across these languages, increasing the number of languages never leads to better performance. These findings suggest that increasing the number of training languages does not necessarily enhance multilingual PRMs. A key reason for this is the fixed amount of training data: as the number of languages grows, the training examples per language decrease. This reduction hinders sufficient training for seen languages and negatively impacts cross-lingual transfer to unseen languages.

\subsection{How Much English Data Do We Need for Multilingual PRMs?}
\label{sec:results_english_percentage}

\begin{figure}[t]
    \centering
    \includegraphics[scale=0.7]{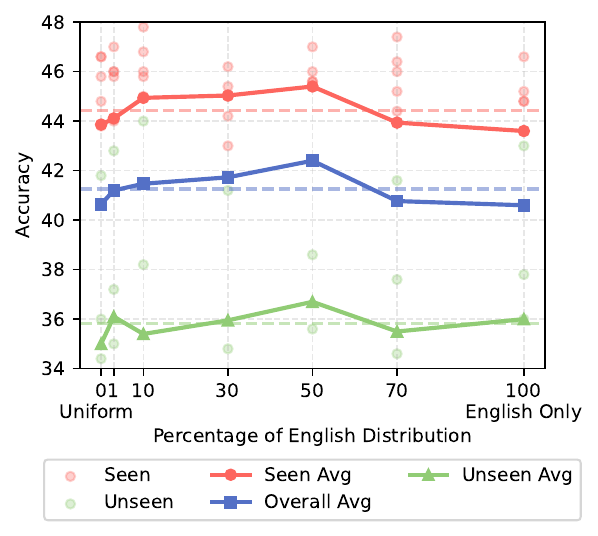}
    \caption{
    Best-of-N sampling performance of \llama 
    with PRMs finetuned on a training set where P\% of the data is in English and (100 - P)\% is uniformly distributed across six other languages. Each tick on the X-axis represents a specific tuning set configuration. 
    The dash lines in {\color{NavyBlue}blue}, {\color{RubineRed} red}, and {\color{YellowGreen}green}, indicate the average scores of all the languages, the seen languages, and the unseen languages, respectively.
}
    \label{fig:english-percentage}
\end{figure}

While multilingual training with equal number of training examples in each language (\mix) generally improves performance compared to English-only training (\en), we observe some exceptions on certain languages, as shown in \autoref{tab:mix-mono-en_math}. This observation prompts us to investigate how varying the number of English examples can affect the multilingual PRMs.

\paragraph{Setup}

To explore this, we create data mixtures with varying percentages of English examples ($P\%$), with the remaining $(100 - P)\%$ examples evenly distributed among six languages: German, Spanish, French, Russian, Swahili, and Chinese. 
Each PRM trained on these mixtures is then evaluated across all 11 languages.

\paragraph{Moderate amount of English data can lead to better multilingual PRMs.}
As shown in \autoref{fig:english-percentage}, incorporating a small amount of English data into the training mixture can lead to notable performance improvements across languages. Specifically, even as little as 1\% of English examples significantly enhances performance, particularly for unseen languages. Interestingly, the majority of performance gains occur when English data constitutes less than 50\% of the training mixture. However, when the proportion of English data exceeds 50\%, performance begins to decline slightly across languages. Furthermore, training on 70\% English data outperforms training solely on English (100\%), suggesting that retaining some multilingual data introduces valuable variation and enhances the generalization capacity of multilingual PRMs. These findings indicate that as the proportion of English data increases, the PRMs may not be adequately trained on other seen languages, and unseen languages may benefit less from cross-lingual transfer. This highlights the importance of maintaining diverse and balanced language representation in multilingual training for optimal performance.

%% file: 6_analysis.tex
\section{Analysis}
\label{sec:analysis}

In this section, we present a comprehensive analysis of our multilingual PRM, focusing on five critical aspects: error positions (\autoref{sec:analysis_error}), number of solutions (\autoref{sec:analysis_samplecounts}), integration of LoRA with PRM (\autoref{sec:analysis_lora}), comparative evaluation with multilingual ORM (see \autoref{sec:analysis_prmorm}), and implement PPO with multilingual PRM (see \autoref{sec:analysis_ppo}).

\subsection{Which Steps Are More Prone to Errors?}
\label{sec:analysis_error}

\begin{figure}[t]
    \centering
    \includegraphics[scale=0.50]{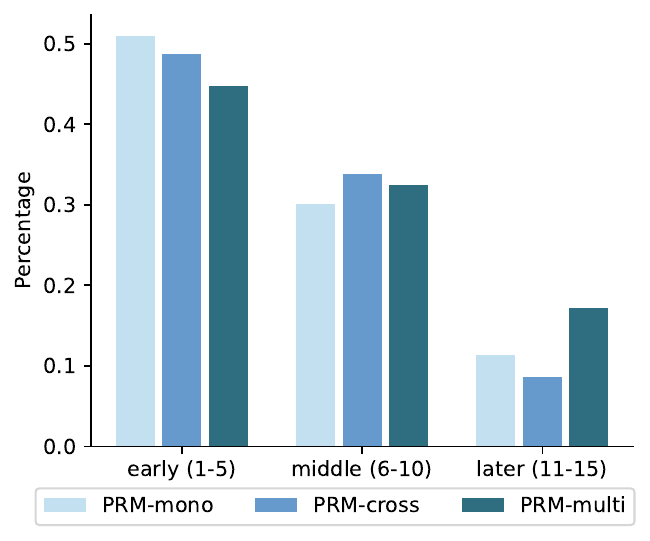}
    \caption{Percentage distribution of the first error positions corresponding to the step in the reasoning on the PRM800K testset.}
    \label{fig:error_ru}
\end{figure}

\begin{figure}[t]
    \centering
    \includegraphics[scale=0.55]{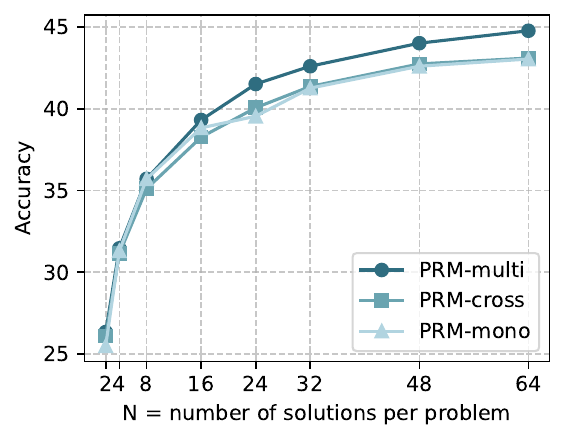}
    \caption{Best-of-N sampling performance of \llama using different verification strategies across distinct numbers of solutions on \mathset.}
    \label{fig:sample-numbers}
\end{figure}

PRMs provide fine-grained feedback on each intermediate step of a model's chain-of-thought reasoning process. Errors at intermediate steps can propagate through the reasoning chain, ultimately affecting the final answer. Therefore, we investigate the earliest errors made by PRMs during the reasoning process, following \citet{processbench}.

\input{tables/baseline}

\paragraph{Setup}
We select a subset of instances from the PRM800K Russian test set where the final answers made by \mono, \en, and \mix are incorrect. For these instances, we identify the first occurrences of incorrect predictions from these PRMs. 
We classify the first error positions into three groups: \textit{early} (steps 1 to 5), \textit{middle} (steps 6 to 10), and \textit{later} (steps 11 to 15).

\paragraph{Multilingual PRMs produce fewer errors at early steps.}

The distribution of the earliest error positions, visualized in \autoref{fig:error_ru}, reveals a clear distinction between the three PRM configurations. In both \mono and \en, a significant proportion of errors occurs within the early steps. In contrast, \mix demonstrates fewer errors within this range and exhibits a slightly higher number of errors in later steps. These observations suggest that \mix may be less prone to error propagation in the reasoning process, enabling it to maintain a more reliable reasoning trajectory. Consequently, \mix can effectively achieve better overall performance.

\subsection{Do More Candidates Drive Better Performance?}
\label{sec:analysis_samplecounts}
Recent research suggests that providing more candidate solutions can significantly boost the performance of PRM \citep{openr,shepherd}. To explore if this applies in multilingual settings, we examine the impact of varying the number of candidates on \mono, \en, and \mix.

\paragraph{Setup}
We conduct experiments on the \mathset benchmark using the \llama generator to compare the performance of \mix, \en, and \mono. For each approach, we vary the number of candidates N from 2 to 64. This allows us to assess how the number of candidate solutions influences performance across different PRM strategies in a multilingual context.

\paragraph{Multilingual PRMs yield better performance with more candidate solutions.}

\autoref{fig:sample-numbers} illustrates that \mix consistently outperforms both \en and \mono, with its advantage growing more pronounced as the number of candidates (N) increases. This finding underscores the scalability of multilingual PRM in diverse linguistic scenarios.
Overall, these observations reinforce the conclusion that multilingual PRM not only maintains superior performance but also scales well as more candidates are introduced.

\subsection{Are Multilingual PRMs Compatible with Parameter-Efficient Finetuning?}
\label{sec:analysis_lora}

Recent research has demonstrated the effectiveness of parameter-efficient finetuning (PEFT) across a variety of tasks \citep{peft1,peft2}. Therefore, we explore whether the PEFT approaches, such as LoRA \citep{lora}, also perform well on multilingual PRMs.

\paragraph{Setup}
To investigate this question, we employ LoRA on the key, query, and value attention matrices. Specifically, we use a rank of 8 and a dropout rate of 0.05 for both multilingual and cross-lingual PRMs. We train for three epochs with a batch size of 64 and a learning rate of $1e^{-5}$.

\paragraph{LoRA is computationally efficient, but not as good as its fully-finetuning counterpart in multilingual PRMs.}

\begin{figure}[t]
    \centering
    \includegraphics[scale=0.6]{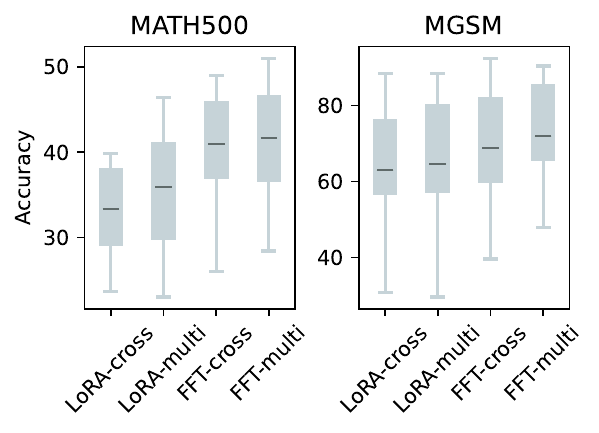}
    \caption{Comparison between parameter-efficient finetuning (LoRA) PRM and fully fine-tuning (FFT) PRM with \llama generator.}
    \label{fig:lora-fft}
\end{figure}

\autoref{fig:lora-fft} demonstrates that fully fine-tuning (FFT) consistently outperforms LoRA in both cross-lingual and multilingual settings. The performance gap becomes larger on the \mathset dataset, which contains more complex questions compared to \mgsmset, suggesting that FFT is better suited for tasks requiring deeper reasoning and understanding. These findings align with prior research, which indicates that while PEFT methods may fall short of FFT when tasks demand higher complexity or reasoning capabilities \citep{lora-fall}. Interestingly, although LoRA-based methods generally lag behind FFT, multilingual LoRA achieves stronger results than cross-lingual LoRA. This highlights the benefits of leveraging multilingual data during parameter-efficient fine-tuning, as multilingual data likely provides richer data diversity and linguistic coverage.

\subsection{Does PRM Surpass ORM in the Multilingual Scenario?}
\label{sec:analysis_prmorm}

In this section, we explore whether PRM also outperforms Outcome Reward Model (ORM) and self-consistency (SC) in multilingual settings.

\paragraph{Setup}

Following \citet{prm800k,shepherd}, we evaluate the performance of \mix by comparing it with other \textit{verifier} methods, including: Direct prediction (\baseline), Self-consistency (majority voting) (\scmethod), and ORM. 
Specifically, we train a multilingual ORM using uniform example budgets across seven seen languages. Then we assess the performance of verifiers on seven seen languages as well as on four additional unseen languages.

\paragraph{Multilingual \prm outperforms \scmethod and \orm across all languages and generators.}

The results presented in \autoref{tab:baseline-sc-orm-prm} confirm that \prm consistently achieves higher accuracy on two benchmarks across multiple languages. Specifically, when using the \llama as the generator, \prm improves average accuracy by +19.64 points on the \mathset dataset and by +15.75 points on the \mgsmset dataset in terms of $\avgall$, compared to the \baseline of direct prediction. These substantial gains suggest \prm's potential to enhance reasoning performance in a multilingual setting. Furthermore, \prm also surpasses both \scmethod and \orm. For example, \prm exceeds SC and \orm by margins of up to +8.80 and +6.73 points on \mgsmset, respectively, when using \llama as the generator. Additionally, \prm demonstrates performance improvements for both seen and unseen languages. With the \deepseek generator on \mgsmset, \prm achieves respective gains of +17.49 and +31.20 for the seen and unseen language sets, compared to the \baseline.

\subsection{Can Multilingual PRM Enhance LLMs?}
\label{sec:analysis_ppo}
\input{tables/ppo}

In this section, we demonstrate that the multilingual PRM can be used as the reward model for finetuning the LLMs under a RL paradigm.

\paragraph{Setup}
We design experiments to improve \llama using RL where we adopt the PPO strategy \citep{ppo} on the MetaMathQA training set \citep{metamath}. We then evaluate the resulting policy models on \mgsmset using top-1 accuracy in a zero-shot setting. Due to the computational constraints, we only generate one response during the fine-tuning process.

\paragraph{Reinforcement learning with multilingual PRM further improves the performance of LLMs.}

The results shown in \autoref{tab:ppo} indicate that step-by-step PPO with \mix (\model{PPO-PRM}) consistently outperforms a standard supervised fine-tuned \model{Baseline} and PPO with ORM (\model{PPO-ORM}). \llama with PPO-PRM achieves average boosts of +1.86 and +2.19 across 11 languages, compared to \model{Baseline} and \model{PPO-ORM}, respectively. These findings highlight the importance of fine-grained multilingual rewards. These gains demonstrate that process rewards can refine policy decisions for both reasoning steps and final outputs with RL. \newrevision{More results are in \autoref{sec:appendix_checkpoint}.}

%% file: tables/baseline.tex
\begin{table*}[t] \small
\centering
\setlength{\tabcolsep}{7pt}
\begin{tabular}{lccccccccc}
\toprule
\multicolumn{10}{c}{\mathset} \\ \midrule
 & \multicolumn{3}{c}{\textsc{Mistral}}  & \multicolumn{3}{c}{\textsc{Llama} }   & \multicolumn{3}{c}{\textsc{DeepSeek}} \\ \cmidrule(rl){2-4} \cmidrule(rl){5-7} \cmidrule(rl){8-10} 
\textit{Verifier} &  $\avgall$    & $\avgseen$     & $\avgunseen$ &  $\avgall$    & $\avgseen$     & $\avgunseen$ &  $\avgall$    & $\avgseen$     & $\avgunseen$ \\\midrule
\baseline         & 22.11     & 24.34     & 18.20    & 22.07    & 24.34     & 18.10    &  26.38    & 32.48    & 15.70   \\
\scmethod & 29.20          & 31.80          & 24.65          & 30.60          & 33.31          & 25.85          & 44.96          & 49.29          & 37.40          \\
\orm              & 39.54          & 42.63          & \textbf{34.25} & 40.49          & 43.14          & 35.85          & 50.96          & 55.54          & 42.95          \\
\mix              & \textbf{39.55} & \textbf{43.11} & 33.30          & \textbf{41.71} & \textbf{44.77} & \textbf{36.35} & \textbf{51.25} & \textbf{55.57} & \textbf{43.70} \\ \midrule

\multicolumn{10}{c}{\mgsmset} \\ \midrule
 & \multicolumn{3}{c}{\textsc{Mistral}}  & \multicolumn{3}{c}{\textsc{Llama} }   & \multicolumn{3}{c}{\textsc{DeepSeek}} \\ \cmidrule(rl){2-4} \cmidrule(rl){5-7} \cmidrule(rl){8-10} 
\textit{Verifier} &  $\avgall$    & $\avgseen$     & $\avgunseen$ &  $\avgall$    & $\avgseen$     & $\avgunseen$ &  $\avgall$    & $\avgseen$     & $\avgunseen$ \\\midrule
\baseline         &  49.63    &  61.65   & 28.60     &  56.18    &  64.23  & 42.10   & 52.95    & 63.02  & 35.30    \\
\scmethod & 56.51          & 69.37          & 34.00          & 63.13          & 74.57          & 43.10          & 70.76          & 75.37          & 62.70          \\
\orm              & 64.84          & 76.40          & 44.60          & 65.20          & 77.43          & 43.80          & 74.44          & 79.00          & 66.45          \\
\mix              & \textbf{65.45} & \textbf{77.09} & \textbf{45.10} & \textbf{71.93} & \textbf{82.00} & \textbf{54.30} & \textbf{75.42} & \textbf{80.51} & \textbf{66.50} \\
\bottomrule
\end{tabular}
\caption{\label{tab:baseline-sc-orm-prm} Multilingual PRMs' best-of-N (N = 64) sampling performance on \mathset and \mgsmset with three generators: \mistral, \llama, and \deepseek. We use \qwen to finetune the \orm and \mix. $\avgall$, $\avgseen$, and $\avgunseen$ indicate the macro-average of results across all the languages, the seen languages, and the unseen languages, respectively.}
\end{table*}

%% file: tables/ppo.tex
\begin{table}[t] \small
\centering
\setlength{\tabcolsep}{4pt}
\begin{tabular}{lccc}
\toprule
&  \baseline & \textsc{PPO-ORM}    & \textsc{PPO-PRM}       \\ \midrule
English  & 78.40    & 80.40   & \textbf{82.40}   \\
German  & 68.80    & 64.00   & \textbf{68.80}   \\
Spanish  & 72.00    & 71.20   & \textbf{76.00}  \\
French  & 67.60    & 68.00   & \textbf{71.60}   \\
Russian  & 69.60    & 68.40   & \textbf{71.20}   \\
Swahili  & 33.60    & 38.80   & \textbf{41.20}   \\
Chinese  & 59.60    & \textbf{64.00}   & 62.80   \\ \hdashline
Japanese  & 48.80    & 46.80   & \textbf{49.20}   \\
Bengali  & \textbf{45.20}    & 41.20   & 40.40   \\
Telugu  & 17.60    & \textbf{20.40}   & 18.00   \\
Thai  & \textbf{56.80}    & 51.20   & \textbf{56.80}   \\ \midrule
Average  & 56.18    & 55.85   & \textbf{58.04}     \\ 
\bottomrule
\end{tabular}
\caption{\label{tab:ppo} Zero-shot evaluation on \mgsmset for \llama improved via PPO with \mix.}
\end{table}

%% file: 7_conclusion.tex
\section{Conclusion}
\label{sec:conclusion}

Through comprehensive evaluations spanning 11 languages, our work demonstrates that multilingual PRMs significantly enhance the ability to perform complex, multi-step reasoning tasks in various languages, consistently outperforming both monolingual and cross-lingual counterparts. Furthermore, our findings highlight that PRM performance is sensitive to the number of languages and the volume of English training data. The multilingal PRMs also benefit from more candidate responses and model parameters. These results underscore the importance of diverse language training in providing fine-grained rewards and open up promising avenues for multilingual reasoning.

%% file: 8_limitations.tex
\section{Limitations}
\label{sec:limitations}

While we have demonstrated the effectiveness of multilingual PRMs, our study has not comprehensively explored the wide range of reward optimization methods \citep{limit1,limit2}, some of which may not benefit from cross-lingual reward model transfer. Nevertheless, best-of-N and PPO, the two techniques leveraged in this paper, are highly representative of current practices, particularly given the consistently strong performance of best-of-N \citep{limit3,limit1,limit4}.
Furthermore, while our results show that multilingual PRMs outperform both cross-lingual and monolingual PRMs, our experiments are limited to 11 languages. Extending this approach to a broader set of languages and evaluating its impact across diverse linguistic families is an important avenue for future work.

%% file: 10_ackowledgement.tex
\section*{Acknowledgement}

This work is funded by EU Horizon Europe (HE) Research and Innovation programme grant No 101070631, and UK Research and Innovation under the UK HE funding grant No 10039436.

The computations described in this research were performed using the Baskerville Tier 2 HPC service (https://www.baskerville.ac.uk/). Baskerville was funded by the EPSRC and UKRI through the World Class Labs scheme (EP/T022221/1) and the Digital Research Infrastructure programme (EP/W032244/1) and is operated by Advanced Research Computing at the University of Birmingham.

%% file: 9_appendix.tex
\section{Data Statistics}
\label{sec:appendix_statistics}

\input{tables/data-statistics}

The dataset statistics are summarized in \autoref{tab:data_stat}. These include the total number of examples, as well as the maximum, minimum, and average number of reasoning steps in the answers across all examples. For the selection criteria for the six languages, there are two key desiderata for the language selection in our work. Firstly, the examples must be accurately translatable into the target language by MT systems. Secondly, the target language must allow for proper evaluation. With these desiderata in mind, we selected six high-resource languages covered by the MGSM dataset. This choice ensures that the translated data closely aligns with the original English dataset and allows us to focus on comparing model strategies without introducing the added variability that lower-resource language translations might cause. We will clarify this in our future revision.

\section{Translation Details}
\label{sec:appendix_translation}
\newrevision{
Due to imbalanced resources across languages, translation has become a standard method for multilingual research. Recent research has demonstrated that machine-translated datasets are comparable to human-translated ones and can be directly used for training and evaluation \citep{DBLP:conf/emnlp/ChenYGH24,DBLP:journals/corr/abs-2410-08928}.}

\newrevision{
In this study, after translating the English dataset into foreign languages, we use regular expressions to filter out the translated training instances that contain discrepancies in numbers or equations compared to the original English dataset. This ensures the correctness of the mathematical content. For the translated multilingual \mathset test set, we employ two human translators to post-edit the test instances in high-resource languages (de, es, fr, ru, zh, and ja) by correcting inaccurate translations and verifying the consistency of mathematical notations. We pay \$0.05 USD for each example, resulting in a total cost of \$150 USD for post-editing. For the low-resource languages (bn, sw, te, and th) in \mathset, we leverage \model{gpt-4o} to post-edit the translations.
}

\newrevision{
To verify the quality of our translations, we use Google Translate to back-translate the multilingual \mathset and 1,000 random training instances from each training set into English. We then calculate the BLEU score using the original English instances as the reference translation. As shown in \autoref{tab:translation}, the high BLEU scores confirm the quality of the translations in our datasets.
}

\begin{table*}
    \centering
    \begin{tabular}{lcccccccccc}
\toprule 
 &  de &  es&  fr&  ru&  sw&  zh&  ja&  bn& te& th\\ \midrule
Train &  81.2&  88.4&  87.3&  74.0&  87.3&  80.8&  -&  -& -& -\\
Test &  85.9&  91.5&  91.0&  73.0&  90.3&  84.4&  84.3&  65.3& 65.8& 80.9\\
\bottomrule
    \end{tabular}
    \caption{BLEU scores of back-translation examples with using the original English data.}
    \label{tab:translation}
\end{table*}

\section{Training Details}
\label{sec:appendix_training}
We train the PRMs by fine-tuning all parameters of \qwen using the AdamW optimizer with a learning rate of $10^{-5}$ and a batch size of 8. This process is conducted over two epochs on 4 NVIDIA A100 GPUs (80GB). During training, we use a linear learning rate schedule with a warm-up phase that constitutes 10\% of the total training steps.

\section{Cross-lingual Transfer of PRMs}
\label{sec:appendix_crosslingualprm}

\begin{figure}[t]
    \centering
    \includegraphics[scale=0.5]{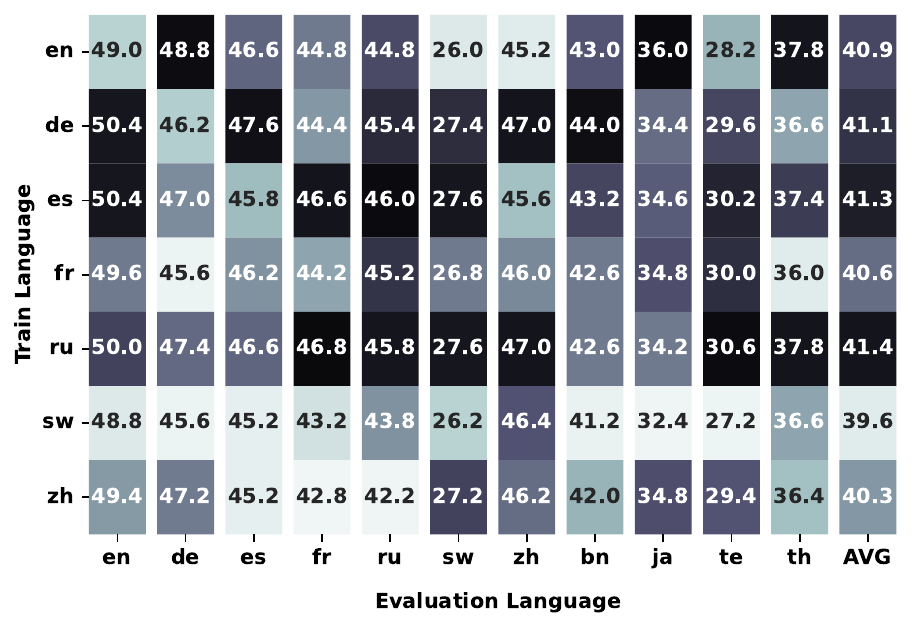}
    \caption{Performance of \mono trained on seven seen languages and evaluated on all 11 languages based on the \mathset with \llama generator.}
    \label{fig:cross-lingual PRM}
\end{figure}

Following \citet{multiorm}, we assess the performance of cross-lingual PRMs to inspect if language similarity like the script or mutual intelligibility might affect the levels of reasoning verification cross-lingual transfer.

\paragraph{Setup}
We train PRMs on monolingual versions of the data in German, Spanish, French, Russian, Swahili, and Chinese, and evaluate their transfer to other languages.

\paragraph{No clear signal indicates that language similarity strongly correlates with cross-lingual transfer.}

We present the cross-lingual transfer results in \autoref{fig:cross-lingual PRM} and observe that there is no clear conclusion regarding the factors that impact cross-lingual transfer. For instance, the PRM trained on Russian data achieves the highest accuracy when evaluating French, Swahili, Chinese, Telugu, and Thai. Notably, these languages neither share the same script nor belong to the same language family as Russian. This observation suggests that linguistic similarity, in terms of script or language family, may not be a decisive factor in cross-lingual transfer. These findings underscore the uncertainty in predicting cross-lingual transfer performance based solely on language similarity. In practice, selecting a diverse set of representative languages for training a multilingual PRM may be a more effective strategy to address this uncertainty and improve performance across a wide range of target languages.

\section{Breakdown Results of \mgsmset for \mono, \en, and \mix}
\label{sec:appendix_mix_mono_en_mgsm}
\input{tables/mix-mono-en-mgsm}

We present the breakdown of results for each language on the \mgsmset in \autoref{tab:mix-mono-en_mgsm}. The results indicate that the \mix consistently outperforms both the \mono and \en models across languages. This observation aligns with the conclusion drawn in \autoref{sec:results_multi_mono}, highlighting the advantages of multilingual training for PRMs.

\section{Results on General-Purpose LLM}
\label{sec:appendix_general}
We provide the results of the general LLM Qwen2.5-7B-Instruct on \mathset in \autoref{tab:general}. It can be observed that the multilingual PRM achieves consistent conclusions when applied to the general LLM.

\begin{table*}[t] \small
\centering
\setlength{\tabcolsep}{3pt}
\begin{tabular}{lcccccccccccccc}
\toprule
\mathset & $\avgall$      & $\avgseen$     & $\avgunseen$   & en             & de             & es             & fr             & ru             & sw             & zh             & ja             & bn             & te             & th             \\ \midrule
\baseline & 36.3 & 37.9 &  33.4 & 44.2& 40.6& 41.6& 40.2& 40.8& 20.8& 37.4& 40.6& 38.2& 20.2& 34.4 \\
\mono & -& 54.2& -& 61.0& 57.6& 58.0& 57.0& 58.2& 31.6& 56.2& -& -& -& \\
\en & 53.2& 57.1& 55.7& 63.6& 60.8& 60.4& 60.4& 61.6& 33.8& 59.4& 60.8& 58.8& 36.8& 56.4 \\
 \mix& 54.0& 58.2& 56.7& 64.8& 61.6& 61.8& 61.2& 62.2& 35.2& 60.6& 61.2& 58.6& 38.2&57.8 \\
\bottomrule
\end{tabular}
\caption{\label{tab:general}Performance on general LLM Qwen2.5-7B-Instruct.}
\end{table*}

\input{tables/ckpt}

\section{Influence of Checkpoint Selection}
\label{sec:appendix_checkpoint}

\newrevision{
We observe a decline in Bengali performance in both ORM and PRM, as shown in \autoref{tab:ppo}. Upon evaluating the performance at each intermediate checkpoint, our analysis indicates that this behavior stems from the PPO training process and the strategy used for selecting the final checkpoint, as illustrated in \autoref{tab:ckpt}. Specifically, since the checkpoint is selected based on the average loss across all languages, the one that minimizes the overall loss does not necessarily yield optimal performance for individual languages. In this case, Bengali appears to follow a distinct learning rate trajectory compared to other languages. We acknowledge this limitation and plan to investigate language-specific adjustments to the training process in future work. 
}

%% file: tables/data-statistics.tex
\begin{table}[t] \small
\centering
\begin{tabular}{lcccc}
\toprule
& \#exam. & max & min & mean\\ \hline
PRM800K trainset & 404K    & 56        & 1         & \phantom{0}6.39       \\
Math-Shepherd   & 445K    & 30        & 1         & \phantom{0}6.23       \\
PRM800K testset  & 5071    & 53        & 1         & 22.11     \\
\bottomrule
\end{tabular}
\caption{
Dataset statistics of the datasets in this work, including number of examples, maximum, minimum, and average number of steps in the answers. 
}
\label{tab:data_stat}
\end{table}

%% file: tables/mix-mono-en-mgsm.tex
\begin{table*}[t] \small
\centering
\setlength{\tabcolsep}{4pt}
\begin{tabular}{lcccccccccccccc}
\toprule
\mgsmset & $\avgall$      & $\avgseen$     & $\avgunseen$   & en             & de             & es             & fr             & ru             & sw             & zh             & ja             & bn             & te             & th             \\ \midrule
\multicolumn{15}{c}{\mistral} \\ \midrule
\mono             & -              & 76.0          & -              & 90.8          & 78.8          & 81.2          & 81.6          & 86.0          & 36.0          & 77.6          & -              & -              & -              & -              \\
\en               & 65.2          & 76.7          & \textbf{45.2} & \textbf{90.8} & \textbf{84.4} & 85.2          & \textbf{82.4} & \textbf{86.8} & 27.2          & \textbf{80.0} & \textbf{76.2} & 43.0          & 7.6           & \textbf{54.0} \\
\mix              & \textbf{65.5} & \textbf{77.1} & 45.1          & 89.2          & 83.2          & \textbf{86.0} & \textbf{82.4} & 86.4          & \textbf{33.2} & 79.2          & 75.6          & \textbf{43.2} & \textbf{8.0}  & 53.6          \\ \midrule
\multicolumn{15}{c}{\llama} \\ \midrule
\mono             & -              & 81.7          & -              & 92.4          & 83.2          & 88.0          & 80.4          & 82.4          & \textbf{62.4} & \textbf{83.2} & -              & -              & -              & -              \\
\en               & 68.8          & 79.3          & 50.6          & \textbf{92.4} & 82.0          & 88.0          & 82.0          & 79.2          & 50.4          & 80.8          & 72.8          & 39.6          & 20.8          & 69.2          \\
\mix              & \textbf{71.9} & \textbf{82.0} & \textbf{54.3} & 90.4          & \textbf{87.6} & \textbf{88.0} & \textbf{83.6} & \textbf{83.2} & 59.6          & 81.6          & \textbf{74.0} & \textbf{48.0} & \textbf{23.6} & \textbf{71.6} \\ \midrule
\multicolumn{15}{c}{\deepseek}  \\ \midrule
\mono             & -              & 80.5          & -              & 96.4          & \textbf{86.4} & 90.4          & 85.2          & 88.0          & \textbf{32.0} & 85.0          & -              & -              & -              & \textbf{-}     \\
\en               & 74.0          & 79.0          & 65.1          & \textbf{96.4} & 86.0          & 91.2          & 85.6          & 87.2          & 18.4          & \textbf{88.4} & 80.0          & 57.6          & 51.6          & 71.2          \\
\mix              & \textbf{75.4} & \textbf{80.5} & \textbf{66.5} & 95.2          & 84.0          & \textbf{92.4} & \textbf{86.4} & \textbf{89.2} & 30.0          & 86.4          & \textbf{80.8} & \textbf{60.8} & \textbf{52.4} & \textbf{72.0} \\
\bottomrule
\end{tabular}
\caption{\label{tab:mix-mono-en_mgsm} Different PRMs' best-of-N sampling (N = 64) performance on \mgsmset with the generator of \mistral, \llama, and \deepseek. $\avgall$, $\avgseen$, and $\avgunseen$ indicate the macro-average of results across all the languages, the seen languages, and the unseen languages, respectively.}
\end{table*}

%% file: tables/ckpt.tex
\begin{table}[H] \small
    \centering
    \begin{tabular}{lccc}
    \toprule
 Checkpoint & Bengali & English & French
\\ \midrule
         \baseline &  45.2&  78.4& 67.6
\\
         Checkpoint-500&  46.8&  80.0& 69.2
\\
         Checkpoint-1150 (final)&  40.4&  82.4& 71.6\\
         \bottomrule
    \end{tabular}
    \caption{The influence of final checkpoint selection strategy during the PPO training process.}
    \label{tab:ckpt}
\end{table}